\documentclass[times, twoside]{zHenriquesLab-StyleBioRxiv}
\usepackage{type1cm} 
\usepackage{lettrine}
\usepackage{xcolor}    
\leadauthor{Hevler} 
\shorttitle{Beyond Sigmoidal Curves: Universal TPP Analysis with Gaussian Processes}

\begin{document}

\title{Thermal Tracks: A Gaussian process-based framework for universal melting curve analysis enabling unconstrained hit identification in thermal proteome profiling experiments}

\author[1,2,3 \Letter]{Johannes F. Hevler}
\author[1,2]{Shivam Verma}
\author[1,2]{Mirat Soijtra}
\author[1,2,3 \Letter]{Carolyn R. Bertozzi}

\affil[1]{Department of Chemistry, School of Humanities and Sciences, Stanford University, Stanford, CA 94305, USA}
\affil[2]{Stanford Chem-H, Stanford University, Stanford, CA 94305, USA}
\affil[3]{Howard Hughes Medical Institute, Stanford, CA 94305, USA}

\maketitle

\begin{abstract}
Thermal Tracks is a Python-based statistical framework for analyzing protein thermal stability data that overcomes key limitations of existing thermal proteome profiling (TPP) workflows. Unlike standard approaches that assume sigmoidal melting curves and are constrained by empirical null distributions (limiting significant hits to $\sim$5\% of data), Thermal Tracks uses Gaussian Process (GP) models with squared-exponential kernels to flexibly model any melting curve shape while generating unbiased null distributions through kernel priors.
This framework is particularly valuable for analyzing proteome-wide perturbations that significantly alter protein thermal stability, such as pathway inhibitions, genetic modifications, or environmental stresses, where conventional TPP methods may miss biologically relevant changes due to their statistical constraints. Furthermore, Thermal Tracks excels at analyzing proteins with unconventional melting profiles, including phase-separating proteins and membrane proteins, which often exhibit complex, non-sigmoidal thermal stability behaviors.
Thermal Tracks is freely available from GitHub and is implemented in Python, providing an accessible and flexible tool for proteome-wide thermal profiling studies.
\end{abstract}


\begin{corrauthor}

Johannes F. Hevler E-mail: jfhevler\at stanford.edu \\
Carolyn R. Bertozzi E-mail: bertozzi\at stanford.edu
\end{corrauthor}

\section*{Introduction}
\begin{figure*}[ht]
    \centering
    \includegraphics[width=1\textwidth]{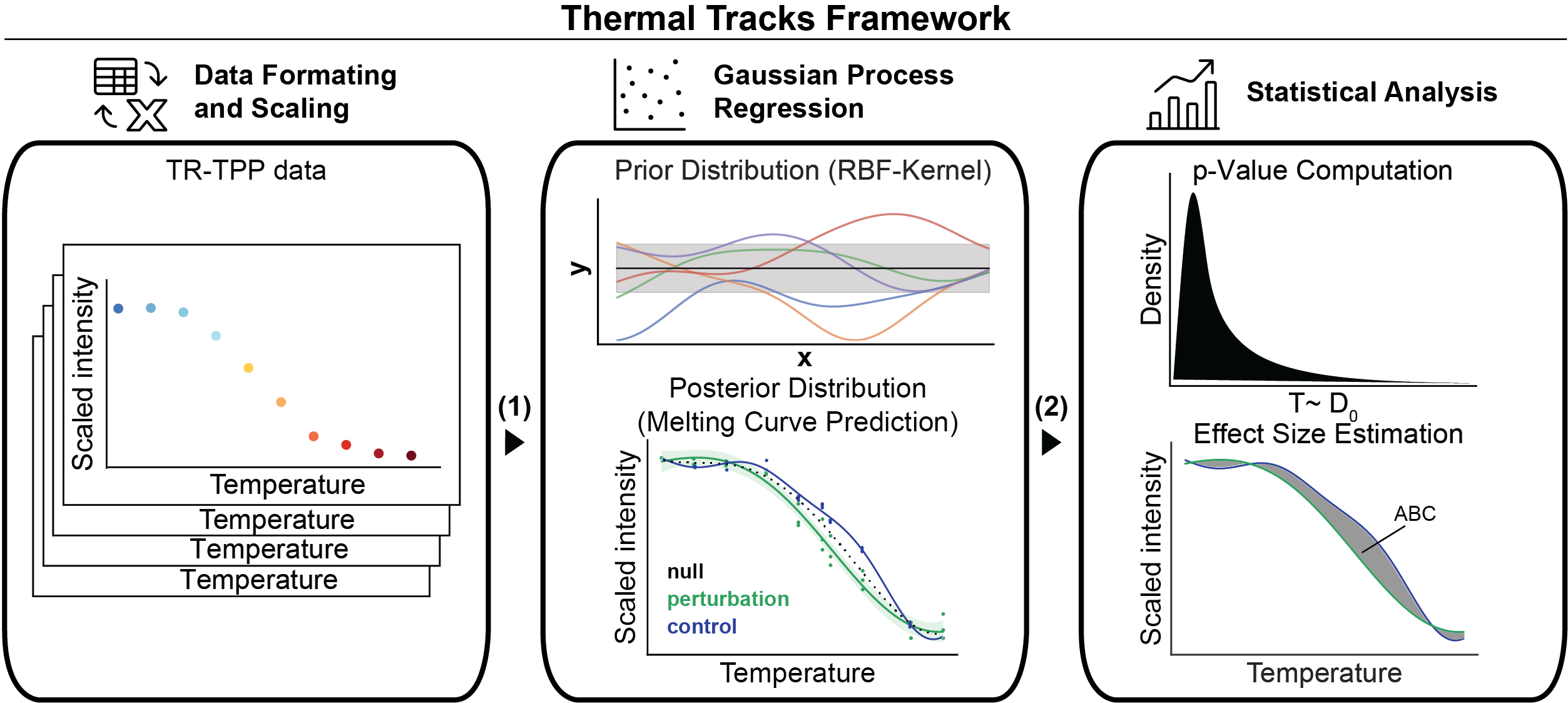}
    \caption{\textbf{Thermal Tracks Framework to overcome current TPP analysis limitations.} The framework fits scaled intensity data from TPP experiments (1) using an RBF kernel (prior distribution) and a Multi-output Gaussian Process (MOGP) model to compare control and treatment conditions. The full model, which includes fits (posterior distribution) for both control (blue) and perturbation (green) conditions, is fitted using Type II maximum likelihood estimation (MLE). Estimated parameters are then incorporated into joint models (black dotted lines), which combine control and treatment conditions. These joint models are used to quantify changes in melting behavior in control vs perturbation by approximating the null distribution of the Thermal Tracks test statistic $\Lambda$ (2). The observed statistics $\Lambda_{\text{control vs. perturbation}}$ are compared to this null distribution approximation to compute empirical p-values. Further, an effect size is determined based on the predicted fits from the full model.}
    \label{Fig1}
\end{figure*}
\lettrine[lines=2,lraise=0.05,findent=0.1em, nindent=0em]{T}hermal Proteome Profiling (TPP) is a powerful proteomics technique that monitors protein thermal stability across the entire proteome by combining the cellular thermal shift assay (CETSA) \cite{RN1} with multiplexed quantitative mass spectrometry using Tandem Mass Tag reagents (TMT) \cite{RN2}. TPP is based on the principle that upon heat-induced stress, proteins denature and become insoluble at characteristic temperatures. By monitoring the remaining soluble protein fractions across a temperature gradient using mass spectrometry, TPP generates thermal stability profiles that are highly informative of protein context and can be influenced by interactions with small molecules (drugs, metabolites), nucleic acids, other proteins, and post-translational modifications \cite{RN7,RN3,RN5,RN6,RN4}. This approach enables investigation of how cellular perturbations influence protein thermal stability and molecular interactions within their native cellular environment.\\
To assess perturbation effects on protein thermal stability, TPP experiments compare melting curves between conditions. The originally introduced data analysis workflow by Franken et al. \cite{RN8}, involves summarizing each curve into a single value, the melting point (Tm), defined as the temperature at which half-maximum relative abundance is observed. Statistical significance is assessed using replicates and hypothesis testing, such as t-tests or z-tests. While important findings have been implicated through comparing melting point differences, it tends to produce high rates of false negatives due to its reliance on sigmoidal assumptions and loss of curve information. To tackle these drawbacks, a method for nonparametric analysis of response curves (NPARC) \cite{RN9} has been introduced. In contrast to summarizing parameters, NPARC employs techniques from functional data analysis \cite{RN10}, enabling comparison of whole melting curves. In brief, the method assumes a sigmoid model for the data and subsequently performs an analysis of variance (ANOVA). Given that TPP data typically involves measuring melting curves for numerous proteins per experiment, the appropriate null distribution can be directly estimated from the data. NPARC enabled the analysis of thousands more proteins than the original Tm-centric approach, significantly enhancing statistical power. However, like the Tm centric approach, this method still relies on a (parametric) sigmoid model and assumes that most observations are derived from the null distribution, which can be problematic when the protein of interest has a non-sigmoidal melting behavior, and a perturbation of interest affects a substantial number of proteins. Moreover, both methods lack uncertainty quantification for the melting curves and the key model parameters.\\
Recent Bayesian approaches have begun addressing the constraints observed for the Tm centric and NPARC approach. Fang et al. introduced Gaussian Process (GP) priors to model non-parametric curve components \cite{RN11}, while Savitski's team developed GPMelt \cite{RN12}, a hierarchical GP framework that relaxes sigmoidal assumptions entirely. Building on these advances, we developed Thermal Tracks to overcome the fundamental limitations of existing TPP analysis workflows: the sigmoidal assumption that fails for proteins with unconventional melting behaviors (such as membrane proteins and phase-separating proteins), and the empirical null distribution that artificially limits hit identification in proteome-wide perturbations.

\section*{Thermal Tracks package}
Like GPMelt developed by Savitski and colleagues, Thermal Tracks removes the sigmoidal assumption often applied to protein melting data while allowing comparison of protein melting behaviors across two conditions (control and perturbation). However, unlike GPMelt's hierarchical approach, Thermal Tracks employs a streamlined architecture without deeper model hierarchies. This design choice reflects that protein-level analysis exhibits less replicate-to-replicate variability compared to peptide-level predictions \cite{RN12}, making hierarchical sub-levels provide limited benefits while potentially increasing computational costs. Additionally, we prioritized creating a user-friendly, computationally efficient workflow implemented in Jupyter notebooks that require no command-line expertise and runs on standard desktop computers without specialized hardware or GPU requirements.\\
Thermal Tracks integrates seamlessly with the Python scientific computing ecosystem, leveraging GPyTorch and PyTorch for efficient Gaussian Process implementation. Because it is implemented in Python with Jupyter notebook functionality, the workflow can be easily extended and customized, enabling direct integration with popular packages, such as pandas for data manipulation and formatting, matplotlib for visualization, and scikit-learn for additional machine learning analyses.
\section*{Thermal Tracks workflow}
In the Thermal Tracks framework (\textbf{Fig. 1}), protein abundances observed across temperature ranges and replicates are modeled using independent Gaussian Process (GP) models for each condition, which are then combined into a multi-output GP model. This approach eliminates manual model loops during fitting and prediction, offering a streamlined and efficient process. We employ the squared-exponential (RBF) kernel to infer underlying melting curves:
\begin{equation}
k_{\mathrm{RBF}}(x_1, x_2) = \exp\left(-\frac{1}{2} (x_1 - x_2)^{\top} \Theta^{-2} (x_1 - x_2)\right)
\end{equation}
where $x_1$ and $x_2$ represent input vectors for observed data points. The matrix $\Theta^{-2}$ contains inverse squared length-scale parameters controlling kernel sensitivity to input variations, allowing flexible adaptation to melting curve shapes. 
Shorter length-scales capture rapid changes, while longer length-scales reflect smoother transitions.\\
The workflow is fully automated and implemented in Python for execution through Jupyter notebooks on standard computers without requiring high-performance computing. After filtering proteins with fewer than three peptide spectral matches (PSMs) or observed in fewer than two replicates, protein abundance data is median-normalized and min-max scaled for consistency across conditions. GP models are trained via type II maximum likelihood estimation using scaled intensities at each temperature.\\
Following the NPARC methodology, the framework relies on hypothesis testing. Two hypotheses are formulated:
\begin{itemize}
    \item \textbf{Null Hypothesis} ($H_0$): There is no difference in a protein's melting curve across conditions.
    \item \textbf{Alternative Hypothesis} ($H_1$): The melting curves for a protein differ between the two tested conditions.
\end{itemize}
To test these hypotheses, a joint model is constructed by combining the trained models from both conditions (control and perturbation) for each protein. This joint model serves as the null model, allowing for the quantification of differences between the conditions. The marginal log-likelihood (mll) values of the joint model and individual models are compared using a likelihood ratio (LR) test:
\begin{equation}
\mathrm{LR} = -2 \times \left( \mathrm{mll}_{\mathrm{joint}} - \left(\mathrm{mll}_{\mathrm{control}} + \mathrm{mll}_{\mathrm{perturbation}}\right) \right)
\end{equation}
where $\mathrm{mll}$ represents a model's marginal log-likelihood. Additionally, the joint model is used to approximate the null distribution by sampling dynamics from a multivariate normal distribution that characterizes the observation distribution under the joint model, similar to the method outlined in Le Sueur et al. \cite{RN12}. To achieve this, an equal number of values are sampled for each comparison and all values across proteins are combined to form a unified null distribution. The observed statistics for the real dataset (see Equation 2) are then compared to this null distribution to compute empirical $p$-values.\\
In the next phase of the workflow, predictions are made based on the generated Gaussian Process (GP) models. To achieve this, both the model and the likelihood are placed in evaluation mode (\texttt{eval}) to prepare for generating predictions on new test data. When evaluating a test point, denoted as $x^*$ with the corresponding true output $y^*$, the model computes the posterior distribution:
\[
p(f^* \mid x^*, X, y)
\]
where $X$ and $y$ represent the training data. This posterior distribution encapsulates the model's uncertainty and gives a function that describes the relationship between the input data and the latent function we are trying to model. The posterior mean and covariance provide a probabilistic estimate, which in the context of protein melting curves helps quantify how well the model fits the thermal stability profile of a protein under different conditions. For example, predicting the melting behavior of a protein at a particular test temperature $x^*$ yields a posterior distribution over the melting curve at that temperature. The mean of this distribution represents the predicted melting temperature, while the covariance quantifies the uncertainty in that prediction.\\ 
Further, by applying the trained likelihood to the model's predictions, we obtain the posterior predictive distribution:
\[
p(y^* \mid x^*, X, y)
\]
which directly models the probability of observing specific melting behavior at $x^*$. This distribution over $y^*$ reflects the uncertainty in both the model and the measurement process, providing a full probabilistic description of the protein's thermal stability.\\
In the final step of the workflow, all results are summarized and saved. The melting curves and correlation matrices for each protein are plotted, and effect sizes are computed (based on the area between the melting curves for each protein and condition).

\section*{Thermal Tracks Performance}
To showcase how Thermal Tracks can be used to analyze TPP datasets and identify thermal stability changes across different conditions, we applied our framework to three distinct datasets representing different analytical challenges. We first analyzed TPP data from cells treated with Staurosporine \cite{RN2}, a kinase inhibitor with well-characterized targets. Thermal Tracks showed high convergence with NPARC and GPMelt for hit identification (\textbf{Fig. 2a}) and successfully identified known Staurosporine targets (\textbf{Fig. 2b}). Additionally, Thermal Tracks identified slightly more significant hits than established methods (NPARC and GPMelt; \textbf{Fig. 2a}) while producing ideal uniform p-value distributions (\textbf{Fig. 2c}). Notably, NPARC's bimodal p-value distribution potentially suggests that its F-statistics already struggle even with this relatively targeted perturbation affecting primarily kinases, potentially leading to missed biologically relevant hits. GPMelt also displayed non-ideal p-value distributions with elevated frequencies in the middle range, potentially hinting towards an overly stringent statistical behavior. In contrast, Thermal Tracks' uniform p-value distribution demonstrates proper statistical calibration. These findings demonstrate that Thermal Tracks exhibits superior performance in experiments with higher levels of targeted perturbations, achieving comparable identification rates as NPARC while simultaneously detecting additional hits that may represent biologically relevant downstream because of the treatment.\\
Next, we applied Thermal Tracks to a more challenging dataset involving high-concentration ATP addition to cell lysates (ATP TPP dataset \cite{RN4}), which creates proteome-wide thermal stability changes affecting numerous nucleotide-binding proteins, far exceeding the ~5\% threshold that constrains conventional TPP analysis methods. Under these conditions where standard workflows struggle due to their empirical null distribution assumptions, Thermal Tracks outperformed NPARC and matched GPMelt's performance in identifying known ATP binders, demonstrating the framework's capability to handle large-scale perturbations without artificial hit identification constraints (\textbf{Fig. 2d}). Most strikingly, when analyzing the melting behavior of the DNA binding protein (likely phase separating) NUCKS1 (nuclear ubiquitous casein kinase and cyclin-dependent kinase substrate 1) (\textbf{Fig. 2e}) or \textit{E. coli} membrane proteins in the presence or absence of high Mg\textsuperscript{2+} concentrations \cite{RN3} (\textbf{Fig. 2f}), Thermal Tracks successfully captured complex, non-sigmoidal melting behaviors that NPARC failed to model. This enabled identification of biologically meaningful thermal stability changes, demonstrating the framework's unique capability to analyze proteins with unconventional melting profiles, a critical advancement for studying challenging protein classes that exhibit complex thermal behaviors.\\

\begin{figure*}[ht]
    \centering
    \includegraphics[width=1\textwidth]{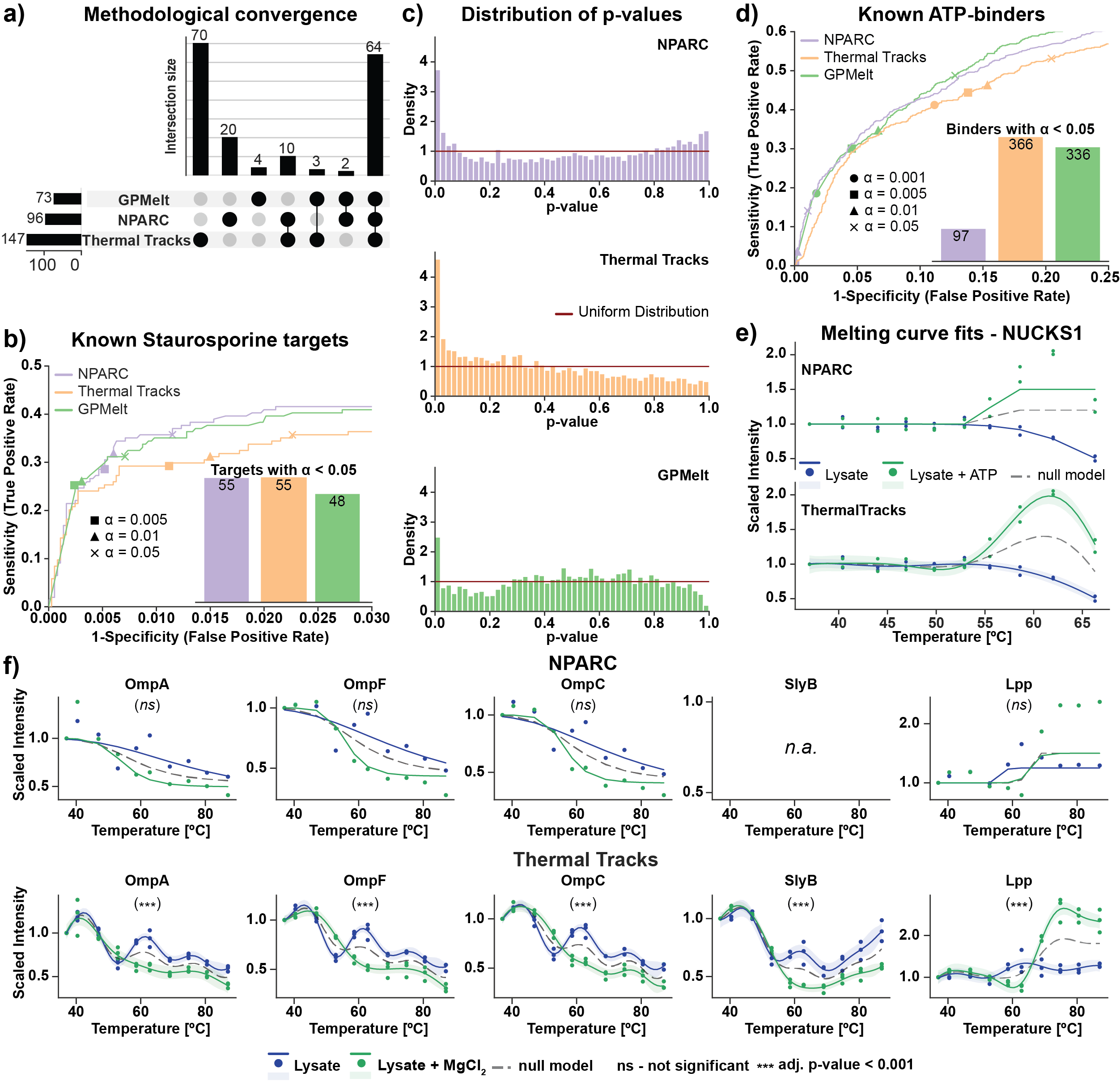}
    \caption{\textbf{Thermal Tracks performance for analyzing TPP data compared to established TPP analysis methods.} \textbf{(a)} UpSet plot showing hit convergence between NPARC, Thermal Tracks, and GPMelt for Staurosporine treatment, with intersection sizes indicating overlap in identified significant proteins. \textbf{(b)} Approximate receiver operator characteristic (ROC) curves comparing NPARC \cite{RN9}, GPMelt \cite{RN12}, and Thermal Tracks on the Staurosporine TPP dataset \cite{RN2}. 176 out of 4505 proteins present kinase activity and are annotated as Staurosporine targets (using annotations provided as supplementary data in Le Sueur et al. \cite{RN12}). The points on the curves correspond to the sensitivity and specificity of NPARC, GPMelt, and Thermal Tracks at an $\alpha$-threshold of $\alpha \in \{0.001, 0.005, 0.01, 0.05\}$ on the BH-adjusted p-values. The ROC curves for the Staurosporine TPP dataset show that Gaussian Process frameworks such as Thermal Tracks perform as well as the gold standard NPARC for capturing affected proteins of more targeted perturbations. Thermal Tracks and NPARC capture 55 known Staurosporine targets with an $\alpha$-threshold of $<0.05$ on the adjusted p-values, compared to 48 for GPMelt. \textbf{(c)} P-value histograms demonstrating statistical calibration: Thermal Tracks produces uniform p-value distributions (ideal), while NPARC shows a bimodal distribution and GPMelt displays elevated middle frequencies, suggesting suboptimal statistical behavior. \textbf{(d)} Approximate ROC curves comparing NPARC \cite{RN9} and GPMelt \cite{RN12}. 753 out of 4772 proteins are annotated as ATP-binding proteins (using annotations provided as supplementary data in Sridharan et al.\cite{RN4}). The points on the curves correspond to the sensitivity and specificity of NPARC, GPMelt, and Thermal Tracks at an $\alpha$-threshold of $\alpha \in \{0.001, 0.005, 0.01, 0.05\}$ on the BH-adjusted p-values. The ROC curves for the ATP TPP dataset show that Gaussian Process frameworks such as Thermal Tracks and GPMelt are more sensitive than NPARC for capturing affected proteins under global perturbations such as ATP ligand binding. Thermal Tracks captures 366 known ATP binders with an $\alpha$-threshold of $<0.05$ on the adjusted p-values, compared to 336 for GPMelt and 97 captured by NPARC. \textbf{(e-f)} Predicted melting curves for the DNA-binding protein NUCKS1 \textbf{(e)} and E. coli membrane proteins \textbf{(f)} based on the ATP TPP dataset and E. coli lysate TPP experiment \cite{RN3} with and without MgCl$_2$, respectively. While NPARC struggles with fitting the non-typical and complex melting curves of NUCKS1 and selected membrane proteins, Thermal Tracks enables accurate fitting of the data, revealing significant differences (*** - BH-adjusted p-values $<0.001$) in melting behaviors across control and perturbation conditions.}
    \label{Fig2}
\end{figure*}\clearpage
\twocolumn
\section*{Conclusions}
We present Thermal Tracks, a Python-based statistical framework implementing Gaussian Process models to overcome fundamental limitations in thermal proteome profiling analysis. In addition to providing a user-friendly Jupyter notebook implementation that runs on standard desktop computers, we provide a method for analyzing proteins with any melting curve shape while enabling unconstrained hit identification through kernel prior-based null distribution modeling. This approach is particularly valuable for proteome-wide perturbations and proteins with unconventional thermal stability behaviors, such as membrane proteins and phase-separating proteins. We demonstrate the superior performance of our framework through comprehensive benchmarking against established methods (NPARC and GPMelt) using three distinct TPP datasets: Thermal Tracks matched existing methods on standard datasets (Staurosporine treatment), outperformed them on proteome-wide perturbations (ATP addition), and uniquely captured complex melting behaviors in membrane proteins (\textit{E. coli} Mg\textsuperscript{2+} treatment) that conventional sigmoidal-based methods failed to detect. All computations were performed on a standard desktop computer using easy-to-use Jupyter notebooks, which are available at the GitHub repository (\href{https://github.com/HevlerJohannes/Thermal-Tracks.git}{https://github.com/HevlerJohannes/Thermal-Tracks.git}) along with details on the analysis and implementation.

\section*{Funding}
This work was supported in part by the National Institutes of Health grant R01CA200423 (C.R.B.), EMBO Postdoctoral Fellowship (ALTF 904-2023; J.F.H.), Howard Hughes Medical Institute Fellowship of the Life Sciences Research Foundation (Award year: 2024; J.F.H) and the Cancer Research Institute Irvington Fellowship (Award year: 2024 \#CRI5766; M.S.).
\newpage
\section*{Bibliography}
\bibliography{bibliography}

\end{document}